\makeatletter\def\graphicscache@inhibit{true}\makeatother

\documentclass[letterpaper, 10 pt, conference]{ieeeconf}  %

\IEEEoverridecommandlockouts                              %

\usepackage[utf8]{inputenc}

\usepackage[hidelinks]{hyperref}
\usepackage[style=ieee,hyperref,natbib=true,backend=bibtex,firstinits,doi=false,%
     mincitenames=1,maxcitenames=2,maxbibnames=99,sorting=none,terseinits=false,hyperref=true]{biblatex}
\bibliography{references.bib}
\renewbibmacro*{bbx:savehash}{}%
\defbibheading{bibliography}[\bibname]{\section*{References}}

\usepackage{url}
\usepackage{graphicx}
\usepackage[usenames,dvipsnames]{xcolor}
\usepackage[draft,inline,nomargin]{fixme}
\fxsetup{theme=color}

\usepackage{amsmath}
\usepackage{amssymb}
\usepackage{textcomp}
\usepackage{siunitx}

\usepackage{booktabs}
\usepackage{threeparttable}
\usepackage{multirow}

\usepackage[capitalize]{cleveref}

\usepackage{tikz}
\usetikzlibrary{arrows}
\usetikzlibrary{positioning,calc}
\usetikzlibrary{decorations.pathreplacing}
\usetikzlibrary{decorations.markings}
\usetikzlibrary{fit}
\usetikzlibrary{shapes.callouts}
\usetikzlibrary{shapes.geometric}
\usetikzlibrary{matrix}

\usepackage{pgfplots}
\pgfplotsset{compat=1.9}
\usepgfplotslibrary{groupplots}
\usepgfplotslibrary{units}
\usepgfplotslibrary{statistics}

\usepackage{blindtext,letltxmacro,xcolor,xparse}
\LetLtxMacro{\blindtextblindtext}{\blindtext}
\LetLtxMacro{\blindtextBlindtext}{\Blindtext}

\RenewDocumentCommand{\blindtext}{O{\value{blindtext}}}{%
  \begingroup\color{gray}\blindtextblindtext[#1]\endgroup
}

\usepackage[export]{adjustbox}

\usepackage{ifthen}
\IfFileExists{graphicscache.sty}{\usepackage{graphicscache}}

\definecolor{teaserbg}{HTML}{ebebeb}

\usepackage{dblfloatfix}

\tikzset{
vr/.style={fill=yellow!40},
manip/.style={fill=blue!20},
loco/.style={fill=green!20}
}

\usepackage[firstpage=true]{background}
\newcommand\copyrighttext{%
\parbox{\textwidth}{
\footnotesize
\centering
\textbf{Accepted final version.} IEEE/RSJ International Conference on Intelligent Robots and Systems (IROS), Prague, Czech Republic, September 2021
}
}
\SetBgContents{\copyrighttext}
\SetBgScale{1}
\SetBgColor{black}
\SetBgAngle{0}
\SetBgOpacity{1}
\SetBgPosition{current page.north}
\SetBgVshift{-0.8cm}
\usepackage[bottom]{footmisc}

\title{\LARGE \bf
NimbRo Avatar: Interactive Immersive Telepresence \\
with Force-Feedback Telemanipulation
}

\author{Max Schwarz$^{*}$, Christian Lenz$^{*}$, Andre Rochow, Michael Schreiber, and Sven Behnke%
\thanks{$^{*}$Equal contribution.}\thanks{All authors are with the Autonomous Intelligent Systems group of University of Bonn, Germany; {\tt schwarz@ais.uni-bonn.de}}%
}

\begin{document}

\maketitle

\begin{abstract}

Robotic avatars promise immersive teleoperation with human-like manipulation
and communication capabilities.
We present such an avatar system, based on the key components of immersive
3D visualization and transparent force-feedback telemanipulation.
Our avatar robot features an anthropomorphic bimanual arm configuration with
dexterous hands. The remote human operator drives the arms and fingers through
an exoskeleton-based operator station, which provides force feedback both
at the wrist and for each finger.
The robot torso is mounted on a holonomic base, providing locomotion capability
in typical indoor scenarios, controlled using a 3D rudder device.
Finally, the robot features a 6D movable head with stereo cameras, which stream
images to a VR HMD worn by the operator. Movement latency is hidden using spherical
rendering. The head also carries a telepresence screen displaying a synthesized image
of the operator with facial animation, which enables direct interaction with
remote persons.
We evaluate our system successfully both in a user study with untrained operators as well as
a longer and more complex integrated mission. We discuss lessons learned from the
trials and possible improvements.

\end{abstract}

\section{Introduction}

Telemanipulation and telepresence are key cornerstones of robotics.
On the one hand, they enable robots to perform tasks which are currently
beyond the capabilities of autonomous perception, planning, and control
methods---the human intellect is still unmatched in its ability to
perceive, plan, and react to unforeseen situations.
On the other hand, they allow humans to work in remote environments
without needing to travel or to expose themselves to potential
dangers, such as in disaster response.
The COVID-19 pandemic has further highlighted the need of
and potentials for teleoperation systems. Telerobotic systems can help
reducing contacts and thus lower the infection risk.
This not only includes medical work, but also helping persons requiring assistance
in their activities of daily life. 

\begin{figure}
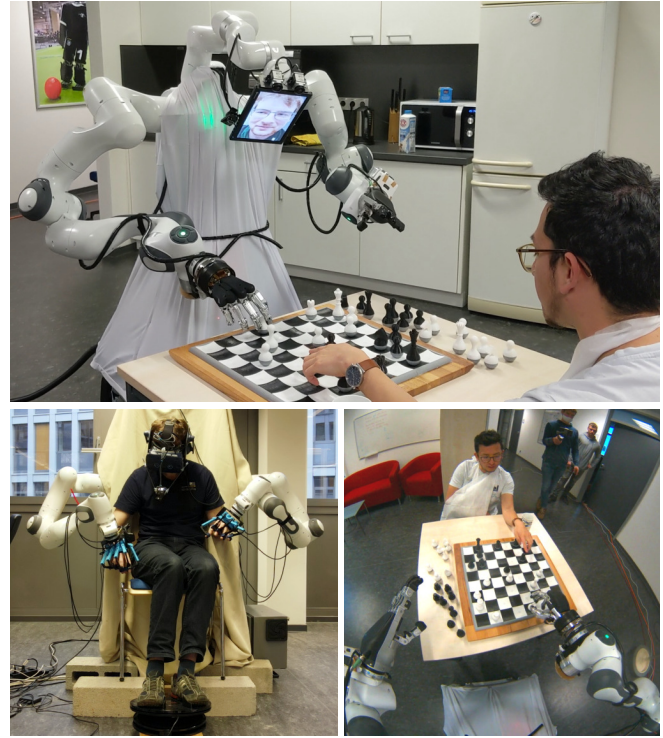

 \centering
 \includegraphics[width=\linewidth,clip,trim=200 20 0 0]{images/teaser/avatar.png}\\
 \vspace{0.3em}
 \includegraphics[height=4.35cm,clip,trim=570 0 520 250]{images/teaser/operator2.png}\hfill
 \includegraphics[height=4.35cm,clip,trim=400 0 400 0]{images/teaser/vr2.png}
  \caption{NimbRo Avatar interacting with a human recipient. Top: Remote environment.
  Bottom left: Human operator. Bottom right: Operator VR view.}
 \label{fig:teaser}
\end{figure}

The ANA Avatar XPRIZE Challenge\footnote{\url{https://www.xprize.org/prizes/avatar}}
fosters development of telerobotic and telepresence systems for these and other
use cases. It focuses on immersiveness and intuitive operation---both
for the remote operator as well as so-called \textit{recipients} interacting
with the robot.
We present our telerobotic system designed for the challenge (see \cref{fig:teaser}), which is
capable of locomotion in human environments,
full 3D immersion, dexterous bimanual manipulation, and interaction with the recipient. Our system
has an approximately humanoid shape, with two arms ending in anthropomorphic hands.
It features a 6D-movable head, carrying cameras and a telepresence screen.
Our operator station facilitates full force feedback to the operator's wrists
and fingers. The operator wears a VR head-mounted display to fully immerse
into the remote environment.
We build the whole system using only off-the-shelf components which allows
for easy replication and maintenance.
~~In summary, our contributions include:

\begin{enumerate}
 \item An anthropomorphic telemanipulation robot (avatar) with advanced manipulation
   and communication capabilities,
 \item a matching operator station, allowing full telepresence and force-feedback telemanipulation,
 \item a real-time operator head animation technique, and
 \item integration into an telerobotic avatar system, including
   a VR low-latency rendering technique and a force-feedback system optimized
   for low-latency operation.
\end{enumerate}

We evaluate our system in lab trials, with both trained and untrained operators to assess
the intuitive control. Furthermore, we demonstrate a complex integrated mission consisting of
six separate tasks---modeled after the ANA Avatar XPRIZE Challenge rules.

\section{Related Work}

Telemanipulation robots are complex systems consisting of many components, which have been the focus of
research both individually as well as on a systems level.

\paragraph{Telemanipulation Systems}

The DARPA Robotics Challenge (DRC) 2015~\citep{krotkov2017darpa} resulted in the development of several
mobile telemanipulation robots, such as DRC-HUBO~\citep{oh2017technical}, CHIMP~\citep{stentz2015chimp}, RoboSimian~\citep{karumanchi2017team}, and our own entry Momaro~\citep{schwarz2017nimbro}.
All these systems demonstrated impressive locomotion and manipulation capabilities under teleoperation,
even with severely constrained communication.
However, the DRC placed no emphasis on intuitiveness of the teleoperation controls or immersion of
the operators. To our knowledge, our team was the only one using a VR HMD and 6D magnetic trackers
to perceive the environment in 3D and to control the robot arms---the rest of the teams relying
entirely on 2D monitors and traditional input devices to control their robots.
All teams, including ours, required highly trained operators familiar with the custom-designed
operator interfaces.
Furthermore, since the DRC was geared towards disaster response, the robots did not feature any
communication capabilities for interacting with remote humans.

In our subsequent work~\citep{klamt2020remote}, we developed the ideas embodied in the Momaro system further.
The resulting Centauro robot is a torque-controlled platform capable of locomotion and dexterous manipulation in rough
terrain. It is controlled by a human sitting in a dedicated operator station, equipped with
an upper body exoskeleton providing force feedback and a VR HMD.
Still, Centauro is focused on disaster response and does not have any communication facilities.

\Citet{Schmaus:RAL18} discuss the results of the METRON SUPVIS Justin space-robotics experiment,
where an astronaut on the ISS controlled the Justin robot on Earth, simulating an orbital robotics
mission. Instead of opting for full immersion and direct control, the authors relied on a 2D tablet
display and higher levels of autonomy, allowing the astronaut to trigger autonomous task skills.

In contrast to the discussed prior works, our avatar system is specifically designed to operate in
human workspaces and to interact with humans. While individual aspects of this problem setting
have been addressed before (and will be discussed below), there exists, to our knowledge,
no integrated system designed for this purpose.

\paragraph{3D VR Televisualization}

Live capture and visualization of the remote scene is typically done using data from RGB or RGB-D cameras.
There are many examples of static and movable stereo cameras on robots, which are directly visualized
in a head-mounted display~\citep{martins2015design,zhu2010head,agarwal2016imitating}. However,
these approaches are limited either by a fixed viewpoint, or considerable camera movement latency,
potentially creating motion sickness. In contrast, our system hides latencies by correcting
for viewpoint changes through spherical rendering~\citep{schwarz2021vr}.

RGB-D sensors allow rendering from free viewpoints~\citep{whitney2020comparing,sun2020new}, removing head movement latency. However, these sensors produce sparse point clouds, which can be difficult to visualize
in a convincing way. Reconstruction-based approaches~\citep{rodehutskors2015intuitive,klamt2020remote,stotko2019vr} address this issue by aggregating
point clouds over time and building dense representations, which can be viewed without movement latency.
They still, however, struggle with many reflective and transparent materials, because the depth sensors
cannot measure them. An additional drawback is that reconstruction-based approaches usually cannot deal
with dynamic scenes---which is an issue when interacting with the environment and human recipients.
In contrast, our method always displays a live stereo RGB stream, which has no difficulties with materials
or dynamic scenes.

\paragraph{Force Feedback}

Teleoperation systems use typically stationary devices to display any force feedback captured by the
remote robot to the human operator ~\cite{hirche2012human, klamt2020remote, abi2018humanoid}. In contrast, wearable haptic devices~\cite{bimbo2017teleoperation}
are usually more lightweight and do not limit the operator's workspace. However, they cannot display
absolute forces to the operator.

Much recent and ongoing research focuses on stable teleoperation systems
in time-delayed scenarios \cite{balachandran2020closing, wang2017adaptive}. Large time
delays for teleoperation in earth-space scenarios are investigated in~\cite{Panzirsch:2020, guanyang2019haptic}.
In our application, we assume smaller distances between the operator station and the avatar robot. Thus, our force feedback controller does not need to handle such high latencies.

\paragraph{Face Animation}

Visualizing facial expressions of persons wearing VR HMDs is a well-known task, also in other contexts.
Usually, IR eye tracking cameras capture eye poses and expressions such as frowns, while a standard camera
captures the unobscured lower part of the face. A special requirement in the Avatar challenge is that
the method needs to be quickly adaptable to a new operator, as less than one hour of setup time is
allotted.

A first category of HMD facial animation methods is based on explicit 3D representations.
\Citet{olszewski2016high} train a neural regressor to output blend shape weights, which deform a face mesh.
On the other hand, Codec Avatars~\citep{chu2020expressive,lombardi2018deep} are an implicit model, trained on
multiple (usually many) images of the operator.

All the mentioned methods require either extensive manual work (3D modeling), complicated capture setups
(3D reconstruction), or long training times, all of which are infeasible in the Avatar challenge.
In contrast, our 2D approach is based on taking a single image of the operator and does not require
any on-site training. However, the resulting quality will be lower than models especially trained or adapted
to the operator at hand.

\section{Operator Station}
\label{sec:operator}

\begin{figure*}
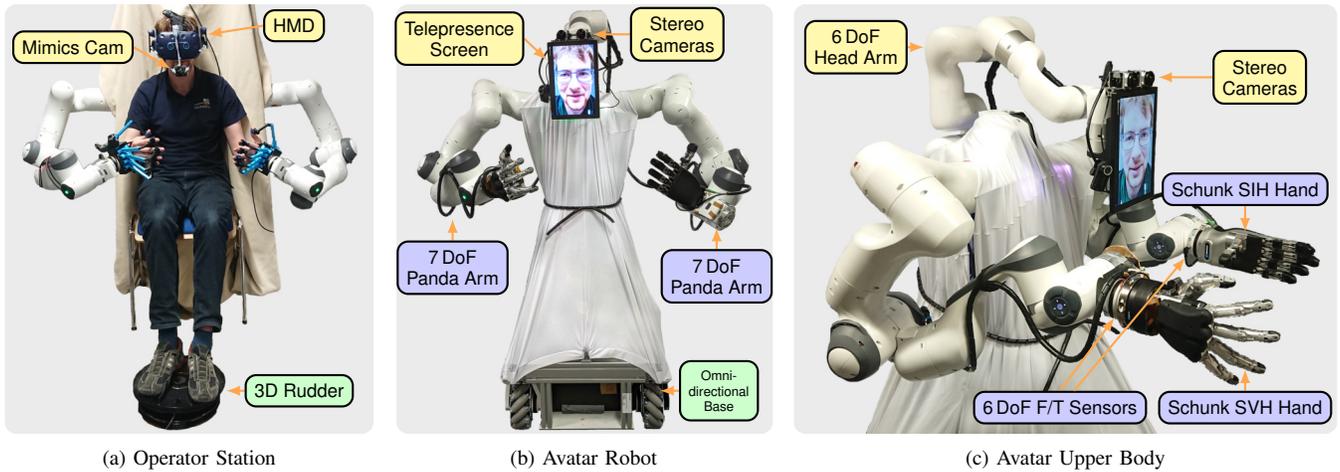

\centering\setlength{\tabcolsep}{0.4em}\tikzset{
  b/.style={fill=teaserbg,rounded corners,inner sep=0pt},
  l/.style={fill=orange!70,rounded corners,align=center,draw=black,font=\sffamily\scriptsize},
  a/.style={draw=orange!70, thick},
}%
 \begin{tabular}{@{}ccc@{}}%
  \begin{tikzpicture}[font=\sffamily]
    \node[b] (img) {\hspace*{.333em}\includegraphics[height=5.7cm,clip,trim=250 1000 550 1700]{images/derOTTO.jpg}\hspace*{.333em}};

    \begin{scope}[x={($ (img.south east) - (img.south west) $ )},y={( $ (img.north west) - (img.south west)$ )}, shift={(img.south west)}]
     \draw[a,latex-] (0.55,0.93) -- ++(0.15,0.02) node[l,anchor=west,vr] {HMD};
     \draw[a,latex-] (0.6,0.1) -- ++(0.05,0) node[l,anchor=west,loco] {3D Rudder};
     \draw[a,latex-] (0.45,0.85) -- ++(-0.1, 0.05) node[l,anchor=east,vr] {Mimics Cam};
    \end{scope}
  \end{tikzpicture} &
  \begin{tikzpicture}
   \node[b] (img) {\hspace*{.333em}\includegraphics[height=5.7cm,clip,trim=0 80 0 20]{images/dieANNA.jpg}\hspace*{.333em}};

   \begin{scope}[x={($ (img.south east) - (img.south west) $ )},y={( $ (img.north west) - (img.south west)$ )}, shift={(img.south west)}]
     \draw[a,latex-] (0.52,0.93) -- ++(0.1,0.0) node[l,anchor=west,vr] {Stereo\\Cameras};
     \draw[a,latex-] (0.4,0.87) -- ++(-0.06,0.05) node[l,anchor=east,vr] {Telepresence\\Screen};
     \draw[a,latex-] (0.7,0.1) -- ++(0.05,0) node[l,anchor=west,font=\sffamily\tiny,loco] {Omni-\\directional\\Base};

     \draw[a,latex-] (0.85,0.48) -- ++(0,-0.05) node[l,anchor=north,manip] {7\,DoF\\ Panda Arm};
     \draw[a,latex-] (0.15,0.5) -- ++(0,-0.05) node[l,anchor=north,manip] {7\,DoF\\ Panda Arm};

    \end{scope}
  \end{tikzpicture} &
  \begin{tikzpicture}
   \node[b] (img) {\hspace*{.333em}\includegraphics[height=5.7cm,clip,trim=400 0 200 200]{images/anna_side.jpg}\hspace*{.333em}};
   \begin{scope}[x={($ (img.south east) - (img.south west) $ )},y={( $ (img.north west) - (img.south west)$ )}, shift={(img.south west)}]
     \draw[a,latex-] (0.67,0.83) -- ++(0.1,0.0) node[l,anchor=west,vr] {Stereo\\Cameras};
     \draw[a,latex-] (0.24,0.9) -- ++(-0.03,0.0) node[l,anchor=east,vr] {6\,DoF\\Head Arm};
     \draw[a,latex-] (0.83,0.15) -- ++(0.0,-0.05) node[l,anchor=north,manip] {Schunk SVH Hand};
     \draw[a,latex-] (0.83,0.47) -- ++(0.0,0.06) node[l,anchor=south,manip] {Schunk SIH Hand};
     \draw[a,latex-] (0.59, 0.27) -- ++(-0.1,-0.17) node[l,anchor=north,manip](tf) {6\,DoF F/T Sensors};
     \draw[a,latex-] (0.72,0.41) -- (tf);
   \end{scope}
  \end{tikzpicture} \\
  \footnotesize (a) Operator Station &
  \footnotesize (b) Avatar Robot &
  \footnotesize (c) Avatar Upper Body \\
 \end{tabular}
 \caption{NimbRo operator station and avatar robot. All components involved in the VR/telepresence system are colored yellow.
The manipulation components are shown in blue.
Finally, the locomotion system is colored green.}
 \label{fig:overview}
\end{figure*}

\begin{figure}
\centering
\scalebox{0.84}{
\begin{tikzpicture}[
 	font=\sffamily\footnotesize,
    every node/.append style={text depth=.2ex},
	box/.style={rectangle, inner sep=0.2, anchor=west, align=center},
	line/.style={black, very thick},
	midway/.append style={font=\sffamily\scriptsize},
	above/.append style={yshift=-0.5ex},
	below/.append style={yshift=0.5ex},
]
\tikzset{every node/.append style={node distance=3.0cm}}
\tikzset{terminal_node/.append style={minimum size=1.0em,minimum height=2em,minimum width={3cm},draw,align=center,rounded corners,fill=yellow!40}}
\tikzset{content_node/.append style={minimum size=1.5em,minimum height=3em,minimum width={width("Search Point")+0.2em},draw,align=center,fill=blue!15!white, rounded corners}}
\tikzset{header_node/.append style={minimum size=1.5em,minimum height=3em,minimum width={width("Search Point")+0.2em},align=center, rounded corners}}
\tikzset{label_node/.append style={near start}}
\tikzset{group_node/.append style={align=center,rounded corners,draw, dashed , inner sep=1em,thick}}
\tikzset{decision_node/.append style={align=center,shape aspect=1.5,minimum width=7.9em,minimum height=5.4em,diamond,draw,fill=yellow!25!white,font=\sffamily\normalsize,node distance=3.9cm}}
\tikzset{l/.style={thick}}

\node(streamcam)[terminal_node,vr] at(0,6) {Mimics Camera};
\node(hmd)[terminal_node,vr] at(0.0,5) {HMD};
\node(anna_model) [terminal_node,manip] at(0.0, 4) {Avatar Model};
\node(otto_arm_controller)[terminal_node,manip] at (0.0, 3){Arm Controller};
\node(ottoFT)[terminal_node,manip] at(0.0,2) {F/T Sensor};
\node(otto_hand_controller)[terminal_node,manip] at(0.0,1) {Hand Controller};
\node(3drudder)[terminal_node,loco] at(0.0,0.0) {3D Rudder};

\begin{scope}[shift={(6.0,0.0)}]
\node(face) [terminal_node,vr] at(0.0, 6) {Telepresence Screen};
\node(head) [terminal_node,vr] at(0.0, 5) {Head Controller};
\node(cams) [terminal_node,vr] at(0.0, 4) {Stereo Cameras};

\node(anna_arm_controller) [terminal_node,manip] at(0.0, 3) {Arm Controller};
\node(annaFT) [terminal_node,manip] at (0.0,2){F/T Sensor};
\node(anna_hand_controller) [terminal_node,manip] at(0.0, 1) {Hand Controller};
\node(base_controller) [terminal_node,loco] at(0.0, 0.0) {Base Controller};

\draw [dashed, gray] (-2.0, -0.7) -- (-2.0, 7.2) -- (2.0, 7.2) -- (2, -0.7) -- (-2.0, -0.7);
\node()[header_node, align = center] at (0.0, 6.7) {\large Avatar Robot};
\end{scope}

\draw [l,-latex] (streamcam.east) -- (face.west) node [midway,above] {Keypoints};
\draw [l,latex-] ([yshift=-0.1cm]hmd.east) -- (cams.west) node [midway,above,sloped] {Images};
\draw [l,-latex] ([yshift=0.1cm]hmd.east) -- ([yshift=-0.1cm]face.west) node [midway,above,sloped] {Eye Tracking};
\draw [l,-latex] (hmd.east) -- (head.west) node [midway,above] {Head Pose};

\draw [l,latex-] ([yshift=-0.1cm]otto_arm_controller.east) -- (annaFT.west) node [midway,above,sloped] {Forces};
\draw [l,-latex] ([yshift=0.1cm]anna_arm_controller.west) -- (anna_model.east) node [midway,above,sloped] {Joint Positions};
\draw [l,-latex] ([xshift=0.1cm]anna_model.south) -- ([xshift=0.1cm]otto_arm_controller.north);
\draw [l,latex-] ([xshift=-0.1cm]anna_model.south) -- ([xshift=-0.1cm]otto_arm_controller.north);
\draw [l,-latex] (ottoFT.north) -- (otto_arm_controller.south);
\draw [l,-latex] (otto_arm_controller.east) -- (anna_arm_controller.west) node [midway,above,sloped] {EEF Pose};

\draw [l,-latex] ([yshift=0.1cm]otto_hand_controller.east) -- ([yshift=0.1cm]anna_hand_controller.west) node [midway,above,sloped] {Joint Positions};
\draw [l,latex-] ([yshift=-0.1cm]otto_hand_controller.east) -- ([yshift=-0.1cm]anna_hand_controller.west) node [midway,below,sloped] {Forces};
\draw [l,-latex] (3drudder.east) -- (base_controller.west) node [midway,above,sloped] {Base Velocity};

\draw [dashed, gray] (-2.0, -0.7) -- (-2.0, 7.2) -- (2.0, 7.2) -- (2.0, -0.7) -- (-2.0, -0.7);
\node()[header_node, align = center] at (0.0, 6.7) {\large Operator Station};

\end{tikzpicture}
}
\caption{Information flow between the system components. Same coloring as in \cref{fig:overview}.}
\label{fig:system}
\end{figure}
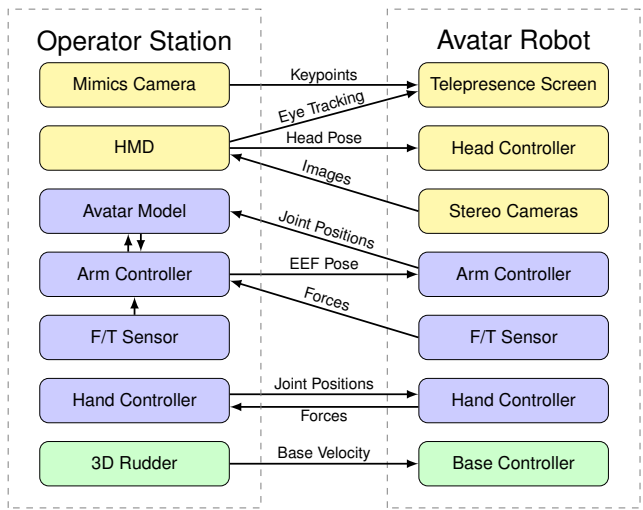

The operator controls the avatar through a dedicated operator station (\cref{fig:overview}a, \cref{fig:system}).
Two Franka Emika Panda 7\,DoF robotic arms are used for
haptic teleoperation.
The arms serve dual purposes: They measure the operator wrist pose precisely and
allow direct inducement of forces at the wrist, thus facilitating force feedback.
The operator hands are connected to the Panda arms through two SenseGlove hand exoskeleton haptic
devices and OnRobot HEX-E 6-axis force-torque sensors between Panda arms and SenseGloves.
The force/torque sensors are used to measure the slightest
hand movement in order to
assist the operator in moving the arm, reducing the felt mass and friction to
a minimum.
The SenseGlove haptic interaction device features 20\,DoF finger joint
position measurements (four per finger) and a 1\,DoF haptic feedback channel
per finger (i.e. when activated the human feels resistance, which prevents
further finger closing movement).

The operator wears an HTC Vive Pro Eye head-mounted display, which
offers 1440$\times$1600 pixels per eye with an update
rate of 90\,Hz and $110^\circ$ diagonal field of view. While other HMDs with
higher resolution and/or FoV exist,
this device offers eye tracking, which is important for visualizing the operator's face on the avatar robot.
The HMD features headphones for audio communication.
We mounted an additional camera (Logitech StreamCam) in front of the operator's lower face to capture
their mimics. The camera is also used for audio capture.

A 3D Rudder foot paddle device is used to capture 3-axis locomotion control
commands. The operator can command translation velocity by pitching and/or rolling their feet,
as well as command yaw velocities by corresponding foot yaw movements.

To ensure safety, the operator station robotic arms are limited in the force they can exert,
causing an immediate shut-down of the system when limits are exceeded. In addition,
a second, supervising person can shut down the system through an E-Stop device.

\subsection{Force-Feedback Control}

Our force feedback controller~\citep{lenz2021teleop} commands joint torques to each Panda arm and reads
the current hand pose to generate the commanded hand pose sent to the avatar
robot. The controller runs with an update rate of 1\,kHz.
To keep the operator and avatar kinematic chains independent, a common control
frame is defined in the middle of
the palm of both the human and robotic hands, i.e. all necessary command and
feedback data are transformed such that they refer to this frame before being
transmitted.
As long as no force feedback is displayed to the operator, we generate a
weightless feeling for moving the arm.
Even though the Panda arm has a quite convenient teach-mode using the internal
gravity compensation when zero torques are commanded, the weightless feeling
can be further improved by using precise external force-torque sensors.
The force-torque measurements are captured with 500\,Hz and smoothed
using a sensor-side low-pass filter with a cutoff frequency of 15\,Hz.

In order to prevent the operator from exceeding joint position or velocity limits
of the Panda arm, we command joint torques pushing the arm away from those limits.
In addition, the avatar arm limits are displayed in a similar way with low latency,
using an operator-side avatar model to predict the avatar arms' movement.

The hand controllers map measured operator finger joint angles to the avatar hands. For
the right Schunk SVH hand, torque feedback in the form of motor currents is
available, which is used to provide per-finger force feedback to the operator.

Any error state of the arms (see \cref{sec:avatar:manip}) is displayed to the operator
as a colored overlay in VR.

\begin{figure}
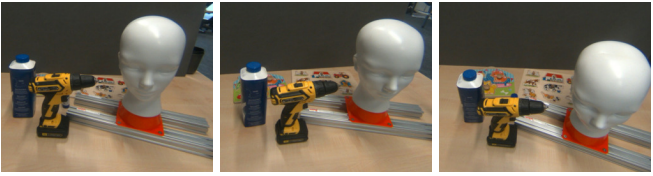

 \centering
 \includegraphics[height=2.25cm,clip,trim=500 200 0 600]{images/perspective/mpv-shot0017.jpg}\hfill
 \includegraphics[height=2.25cm,clip,trim=500 200 0 600]{images/perspective/mpv-shot0011.jpg}\hfill
 \includegraphics[height=2.25cm,clip,trim=500 200 0 600]{images/perspective/mpv-shot0016.jpg}
 \caption{Immersive 3D Visualization: Operator view (cropped) from different head perspectives.}
\end{figure}

\begin{figure}
 \centering
 \begin{tikzpicture}[font=\footnotesize,scale=1.3]
  \begin{scope}[]
   \begin{scope}
    \draw (2,0) arc (0:180:2) -- cycle;
    \clip (2,0) arc (0:180:2) -- cycle;

    \begin{scope}[shift={(-0.8,0.8)}, rotate=-30]
       \draw[fill=green!50]
       (0,0) -- (60:4) arc [radius=4,start angle=60, delta angle=60] -- cycle;
    \end{scope}
    \draw[latex-latex] (15:0.1) -- node[midway,fill=white] {$r$} (15:2);
   \end{scope}
   \draw[fill=white]
       (-0.1,-0.2) rectangle (0.1,0.0)
       (0.0,0.0) -- ++(-0.1,0.1) -- ++(0.2,0.0) -- cycle;

   \begin{scope}[shift={(-0.8,0.8)}, rotate=-30]
   \draw[fill=green]
       (-0.1,-0.2) rectangle (0.1,0.0)
       (0.0,0.0) -- ++(-0.1,0.1) -- ++(0.2,0.0) -- cycle;
   \end{scope}

   \draw[-latex] (0,0) to [in=-60,out=170] node [pos=0.9,right,font=\tiny] {$T^C_V$} (-0.8,0.55);
  \end{scope}
 \end{tikzpicture}
 \caption{%
   Spherical Rendering compensates pose differences between VR eye and robot camera poses (transform $T^C_V$).
   The real camera FoV (green) is projected onto a sphere around the VR camera with radius $r$.
 }
 \label{fig:spherical}
\end{figure}
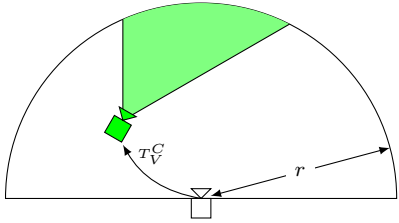

\subsection{3D Visualization}

We took special care to develop an immersive 3D visualization approach, which displays
the environment around the avatar in VR to the operator. Our method achieves both
low-latency response to head movements as well as real-time streaming of dynamic scenes.
State-of-the-art systems can usually only achieve one of these goals.
Our approach uses a 6D movable head on the robot, carrying a stereo camera with human baseline (see \cref{sec:avatar}).
The head mimics operator movements exactly. To compensate for movement latency, which
introduces a pose difference $T^C_V$, we use spherical rendering~\citep{schwarz2021vr}.
In short, the wide-angle camera images are rendered on the operator side as spheres with radius $r=1$\,m, centered on the
camera pose at time of capture. The operator can move their head in VR freely with low latency.
For translations, this causes a small amount of distortion (see \cref{fig:spherical}), but
this is unnoticeable in most situations and is quickly corrected when the robot head arrives at
its target pose.

\section{Avatar Robot}
\label{sec:avatar}

The avatar robot is designed to interact with humans and made-for-humans everyday objects and
 indoor environments and thus
features an anthropomorphic upper body (\cref{fig:overview}b, \cref{fig:system}).
Two 7\,DoF Franka Emika Panda arms are mounted in slightly V-shaped
angle to mimic the human arm configuration. The shoulder height of 125\,cm above
the floor allows convenient manipulation with objects on a table, as well
interaction with both, sitting and standing persons. The shoulder width
of under 90\,cm enables easy navigation through standard doors.

\subsection{Manipulation}
\label{sec:avatar:manip}

\begin{figure}
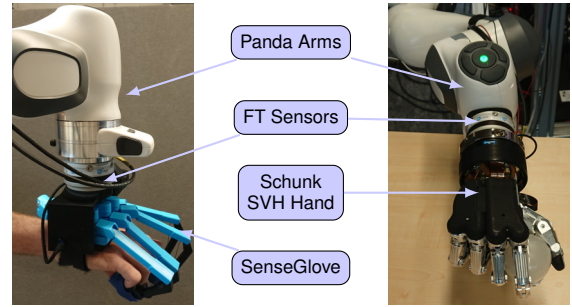

\centering
\begin{tikzpicture}[
 	font=\sffamily\scriptsize,
    every node/.append style={text depth=.2ex},
	box/.style={rectangle, inner sep=0.2, anchor=west, align=center},
	line/.style={black, thick}
]
\tikzset{every node/.append style={node distance=3.0cm}}

\node (otto)[] at (0.0,0.0) {\includegraphics[height=4cm]{images/o_system.png}};
\node (anna)[] at (5.0,0.0) {\includegraphics[height=4cm]{images/a_system.png}};

   \node[manip,rounded corners, draw=black, align=center] (senseGlove) at (2.5,-1.5) {SenseGlove};
   \draw[draw=blue!20, thick, -latex] (senseGlove) -- (1.0,-1.0);

   \node[manip,rounded corners, draw=black, align=center] (panda) at (2.5,1.5) {Panda Arms};
   \draw[draw=blue!20, thick, -latex] (panda) -- (0.3,0.8);
   \draw[draw=blue!20, thick, -latex] (panda) -- (4.8,0.9);

   \node[manip,rounded corners, draw=black, align=center] (ft) at (2.5,0.5) {FT Sensors};
   \draw[draw=blue!20, thick, -latex] (ft) -- (0.0,-0.35);
   \draw[draw=blue!20, thick, -latex] (ft) -- (5.0,0.45);

   \node[manip,rounded corners, draw=black, align=center] (schunk) at (2.5,-0.5) {Schunk\\SVH Hand};
   \draw[draw=blue!20, thick, -latex] (schunk) -- (5.0,-0.5);

   \end{tikzpicture}

 \caption{Haptic interaction system. Operator interface (left) and avatar robotic hand (right).}
 \label{fig:arms}
\end{figure}

The Panda arms have a sufficient payload of 3\,kg, a maximal workspace of
855\,mm and the extra degree of freedom gives some flexibility in the
elbow position while moving the arm. Despite existing torque sensors
in each arm joint, we mounted additional OnRobot HEX-E 6-axis force/torque
sensors at the wrists for more accurate force and torque measurements
close to the robotic hands, since this is the usual location of contact
with the robot's environment (see \cref{fig:arms}). The avatar is equipped with two Schunk hands.
A 20\,DoF Schunk SVH hand is mounted on the right side. The nine actuated
degrees of freedom provide very dexterous manipulation capabilities.
The left arm features a 5\,DoF Schunk SIH hand for simpler but more
force-requiring manipulation tasks. Both hand types thus complement each other.

Software-wise, the arms are driven with a Cartesian Impedance controller
towards the operator's hand pose. The arms feature a safety mechanism
which prevents the arms from exceeding certain joint torque, position, and
velocity limits. In case a software stop occurs (for example when exceeding
torque limits while touching a table) the arm can recover automatically by
smoothly fading between the current and target arm pose.

Any force/torque measured by the
sensor on the wrist is transmitted to the operator side (see \cref{sec:operator}).
Similarly, the hands receive position commands and transmit current measurements
back to the operator station.

\subsection{3D Perception}

Our robot's head is mounted on a UFACTORY xArm 6, providing full 6D
control of the head pose. The robotic arm is capable of moving a 5\,kg payload,
which is more than enough for a pair of cameras
and a small color display for telepresence.
Furthermore, the arm is very slim, which results in a large workspace
while being unobtrusive. Finally, it is capable of fairly high speeds
(180\,${}^\circ$/s per joint, 1\,m/s at the end-effector), thus being able
to match dynamic human head movements.

Two Basler a2A3840-45ucBAS cameras are mounted on the head in a stereo configuration.
The cameras offer 4K video streaming at 45\,Hz and
are paired with C-Mount wide-angle lenses, which provide
more than $180^\circ$ field of view in horizontal direction. We also
experimented with Logitech BRIO webcams with wide-angle converters,
which offer auto-focus but can only provide 30\,Hz at 4K, resulting
in visible stutters with moving objects.
The Basler cameras are configured with a fixed exposure
time (8\,ms) to reduce motion blur to a minimum. For transmission, the raw
camera images are MJPEG-compressed on the onboard GPU~\citep{holub2012ultragrid}.
The entire pipeline achieves 30-40\,ms latency from camera exposure start to
outputting the image to the VR HMD, as measured by the camera-provided timestamps.

\subsection{Telepresence Screen}

\begin{figure}
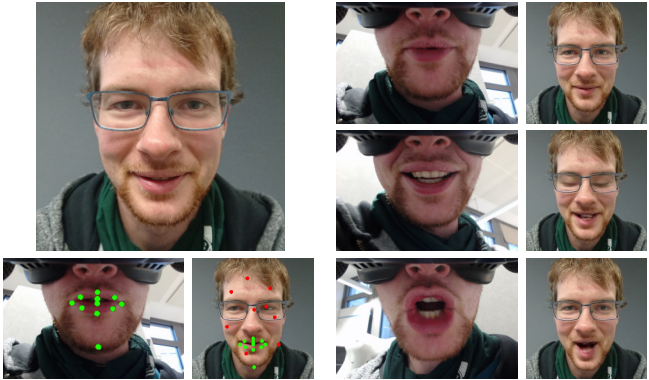

 \centering
 \begin{minipage}{.49\linewidth}\centering
  \includegraphics[height=3.3cm]{images/opvis/chris.jpg}\\[0.1cm]
  
  \includegraphics[height=1.6cm,clip,trim=0 0 300 0]{images/opvis/440_mouthCam_KP_merged.png}
  \includegraphics[height=1.6cm]{images/opvis/440_animated_KP.png}\,
 \end{minipage}\hfill%
 \begin{minipage}{.49\linewidth}\centering
  \includegraphics[height=1.6cm,clip,trim=0 0 300 0]{images/opvis/440_mouthCam.png}
  \includegraphics[height=1.6cm]{images/opvis/440_animated.png}\\[0.1cm]
  
  \includegraphics[height=1.6cm,clip,trim=0 0 300 0]{images/opvis/370_mouthCam.png}
  \includegraphics[height=1.6cm]{images/opvis/370_animated.png}\\[0.1cm]

  \includegraphics[height=1.6cm,clip,trim=0 0 300 0]{images/opvis/2360_mouthCam.png}
  \includegraphics[height=1.6cm]{images/opvis/2360_animated.png}

 \end{minipage}
 
 \caption{Facial animation for the telepresence screen.
   The photograph of the operator (top left) is warped and inpainted using a learned
   network. The network operates on keypoints (bottom left) extracted from the
   lower face camera (green), and head keypoints (red), modified by eye tracking data (not shown).
   The right column
   shows the generated result. Note that the generated image shows a variety
   of mouth and chin poses, as well as eye movements and blinking.}
 \label{fig:opvis}
\end{figure}

For life-like conversations, it is necessary to display the operator's
face with all its expressiveness, which is difficult to achieve since the
operator is wearing an HMD. To address this, we extract facial keypoints from
the lower face camera and the eye tracking data.
We train a keypoint detector network based on an hour glass architecture\footnote{
Network architecture: 128$\times$128 input size, five encoder blocks Conv2d-BN-ReLU-AvgPool,
five decoder blocks Upsample-Conv2d-BN-ReLU. Output is one heat map per keypoint.
} for lower face parts. For training, we crop images from a VoxCeleb2~\citep{VoxCeleb2} sequence and use extracted keypoints~\citep{bulat2017far} as ground truth.

We also learn unsupervised keypoints for the entire face following the method of \citet{Siarohin_2019_NeurIPS}.
The unsupervised keypoints include an eye keypoint controlling gaze direction and blinking and are used to animate a given face in a different facial expression. During training, we sample two images
of the same person, denoted \textit{source} and \textit{target}. We extract both keypoint types from the
target image and ask the generator network to produce the target image, given source image and target keypoints.

During inference, we extract the unsupervised keypoints in the operator photograph and apply
the eye tracking data captured by the HMD on the eye keypoint in pixel space.
We extract the lower face keypoints in the images obtained with the operator cam. 
Unlike in training, we merge the keypoints in the source image operator pose and feed them into a generator network, following \citet{Siarohin_2019_NeurIPS}.
The source image is then adjusted to the target keypoints by the generator network.
We removed keypoint Jacobians from the pipeline, since our keypoint detector does not produce them in the upper face, due to occlusions caused by the VR glasses. %
The generated image is then displayed on the avatar's head in real time.
\Cref{fig:opvis} shows example outputs of the entire pipeline.

For audio communication, the avatar is equipped with a speaker in its torso and a microphone
attached to the head.

\subsection{Locomotion \& Tether}

The robot upper body is mounted on a holonomic platform with four Mecanum
wheels, capable of up to 2.5\,m/s movement, although we usually cap
the operator command at 1.5\,m/s for safety reasons.

Our system currently operates with a communication \& power tether, providing
a 1\,Gbit/s Ethernet connection through which all communication between operator station and avatar robot takes place (see \cref{fig:system}). In its present configuration, the system
consumes around 200\,Mbit/s of bandwidth. Since the stereo 4K camera feed is
responsible for nearly all of the bandwidth, the bandwidth can be freely adapted by
changing video feed resolution and/or frame rate.
The system can be operated with moderate amounts of communication latencies.
When exceeding 50\,ms, the force feedback mechanism will start to become unstable.

\subsection{Safety}

Both subsystems, the operator station and avatar robot, are designed to operate
in contact with humans. Thus, safety mechanisms are highly important.
The Franka Emika Panda arms employed in both systems feature considerable measures
that detect abnormal situations, such as excessive force or speed, and immediately
switch into a safe shutdown state, which holds the current position.

On any communication loss or if the operator station is paused, the avatar arms
remain in their current pose and will
smoothly fade to the new target pose upon receiving new commands.
During operation, the avatar arms are compliant using our impedance controller.

The avatar robot also features a 6\,DoF head arm, which has protective measures similar
to the Panda arms. If the E-Stop is activated, it will stop immediately.
Furthermore, the holonomic base with its four wheels becomes de-energized immediately
when the E-Stop is pressed, allowing the robot to be easily moved by hand.

\section{Evaluation}

\begin{figure}
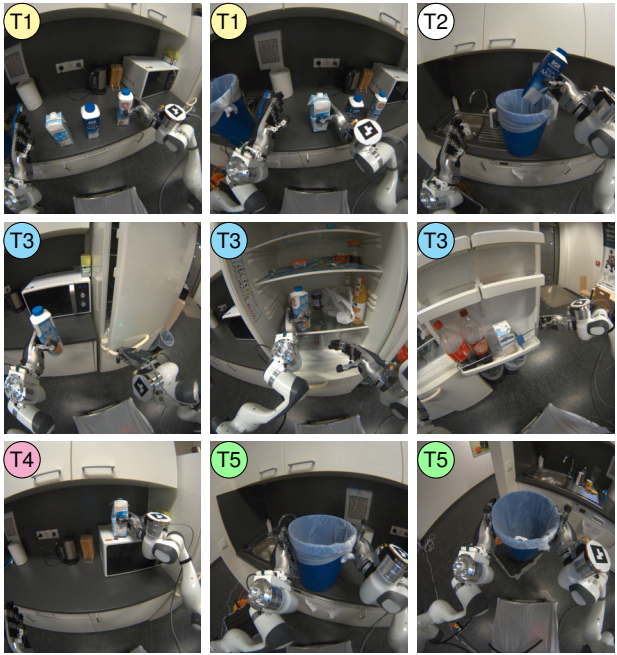

 \centering
 \tikzset{i/.style={inner sep=0cm,anchor=north west}, l/.style={inner sep=1pt,anchor=north west,draw=black,circle,text=black,font=\sffamily\footnotesize,fill=white,shift={(2pt,-2pt)}},
 ll/.style={l, inner sep=0.5pt, font=\sffamily\scriptsize}}

 \tikz{\node[i]{\includegraphics[height=2.8cm,clip,trim=100 0 100 0]{images/userstudy/mpv-shot0001.jpg}}; \node[l,fill=yellow!40]{T1}}
 \tikz{\node[i]{\includegraphics[height=2.8cm,clip,trim=100 0 100 0]{images/userstudy/mpv-shot0010.jpg}}; \node[l,fill=yellow!40]{T1}}
 \tikz{\node[i]{\includegraphics[height=2.8cm,clip,trim=100 0 100 0]{images/userstudy/mpv-shot0002.jpg}}; \node[l]{T2}}
 \\[0.3em]
 \tikz{\node[i]{\includegraphics[height=2.8cm,clip,trim=100 0 100 0]{images/userstudy/mpv-shot0003.jpg}}; \node[l,fill=cyan!40]{T3}}
 \tikz{\node[i]{\includegraphics[height=2.8cm,clip,trim=100 0 100 0]{images/userstudy/mpv-shot0004.jpg}}; \node[l,fill=cyan!40]{T3}}
 \tikz{\node[i]{\includegraphics[height=2.8cm,clip,trim=100 0 100 0]{images/userstudy/mpv-shot0005.jpg}}; \node[l,fill=cyan!40]{T3}}
 \\[0.3em]
 \tikz{\node[i]{\includegraphics[height=2.8cm,clip,trim=100 0 100 0]{images/userstudy/mpv-shot0006.jpg}}; \node[l,fill=magenta!40]{T4}}
 \tikz{\node[i]{\includegraphics[height=2.8cm,clip,trim=100 0 100 0]{images/userstudy/mpv-shot0007.jpg}}; \node[l,fill=green!40]{T5}}
 \tikz{\node[i]{\includegraphics[height=2.8cm,clip,trim=100 0 100 0]{images/userstudy/mpv-shot0008.jpg}}; \node[l,fill=green!40]{T5}}

 \caption{User study tasks. We show cropped versions of the operator left eye VR view. Tasks:
  {\protect \tikz{\protect \node[ll,fill=yellow!40]{1}}} Sorting the cartons by weight.
  {\protect \tikz{\protect \node[ll,fill=white]{2}}} Throwing away the empty carton.
  {\protect \tikz{\protect \node[ll,fill=cyan!40]{3}}} Opening the fridge, putting the half-full carton inside, and closing the fridge.
  {\protect \tikz{\protect \node[ll,fill=magenta!40]{4}}} Storing the full carton on the shelf.
  {\protect \tikz{\protect \node[ll,fill=green!40]{5}}} Grasping the waste basket with both hands and putting it down.
 }
 \label{fig:userstudy:tasks}
\end{figure}

Our system has been evaluated in two larger experiments, focusing on
intuitiveness and immersion in a user study and system capability
in a more complex longer integrated mission.

\begin{figure*}[h]
 \centering
\begin{tikzpicture}[font = \footnotesize, every mark/.append style={mark size=0.5pt}]
 \begin{axis}[
     name=plot,
     boxplot/draw direction=x,
     width=0.53\textwidth,
     height=6cm,
     boxplot={
         draw position={1.8 - 0.325/1 + 1.0*floor((\plotnumofactualtype + 0.001)/1) + 0.2*mod((\plotnumofactualtype + 0.001),1)},
         box extend=0.6,
         average=auto,
         every average/.style={/tikz/mark=x, mark size=1.5, mark options=black},
         every box/.style={draw, line width=0.5pt, fill=.!40!white},
         every median/.style={line width=2.0pt},
         every whisker/.style={dashed},
     },
     ymin=1,
     ymax=12,
     y dir=reverse,
     ytick={1,2,...,13},
     y tick label as interval,
     yticklabels={
\hspace{-1em}1) Were you able to clearly see and hear what was happening in the remote space?,
2) Did you get the necessary haptic feedback to complete the tasks?,
3) Were you able to sense your own position and movement?,
4) Did you feel present in the remote environment?,
5) Was it easy and comforable to use the Avatar System?,
6) Did you feel safe using the Avatar System?,
7) Did you feel the Avatar System was safe for the remote environment?,
8) Was it intuitive to control the arms?,
9) Was it intuitive to control the fingers?,
10) Could you judge depth correctly?,
11) Was the VR experience comfortable for your eyes?
     },
     y tick label style={
         align=center
     },
     xmin=0.75,
     xmax=7.25,
     xtick={1, 2 ,..., 7},
     xticklabels = {1, 2, ..., 7},
     cycle list={{green!50!black}},
     y dir=reverse,
     legend image code/.code={
         \draw [#1, fill=.!40!white] (0cm,-1.5pt) rectangle (0.3cm,1.5pt);
     },
     legend style={
         anchor=north west,
         at={($(0.0,1.0)+(0.2cm,-0.1cm)$)},
     },
     legend cell align={left},
 ]

  \addplot
  table[row sep=\\, y index=0] {
  data\\
  6\\7\\6\\7\\6\\6\\5\\6\\6\\6\\
  };

  \addplot
  table[row sep=\\, y index=0] {
  data\\
  5\\7\\5\\7\\7\\6\\6\\4\\5\\5\\
  };

  \addplot
  table[row sep=\\, y index=0] {
  data\\
  6\\7\\6\\6\\6\\6\\6\\7\\7\\3\\
  };

  \addplot
  table[row sep=\\, y index=0] {
  data\\
  7\\7\\7\\7\\7\\7\\7\\5\\6\\7\\
  };

  \addplot
  table[row sep=\\, y index=0] {
  data\\
  6\\7\\6\\6\\5\\6\\5\\6\\6\\6\\
  };

  \addplot
  table[row sep=\\, y index=0] {
  data\\
  7\\7\\6\\7\\7\\7\\4\\7\\6\\7\\
};

  \addplot
  table[row sep=\\, y index=0] {
  data\\
  7\\7\\4\\7\\5\\7\\4\\3\\6\\6\\
};

  \addplot
  table[row sep=\\, y index=0] {
  data\\
  7\\7\\7\\7\\7\\7\\5\\5\\6\\6\\
};

  \addplot
  table[row sep=\\, y index=0] {
  data\\
  6\\6\\6\\7\\7\\7\\6\\3\\4\\6\\
};

  \addplot
  table[row sep=\\, y index=0] {
  data\\
  6\\7\\7\\7\\7\\7\\5\\3\\6\\7\\
};

  \addplot
  table[row sep=\\, y index=0] {
  data\\
  5\\7\\6\\6\\7\\7\\6\\3\\5\\7\\
};

 \end{axis}

 \draw[-latex] ($(plot.south west)+(0.3cm,0.2cm)$) -- ($(plot.south west)+(1.4cm,0.2cm)$)  node[midway,above,font=\scriptsize,inner sep=1pt] {better};

 \end{tikzpicture}
  \vspace{-1cm}
 \caption{
   Statistical results of our user questionnaire. We show the median, lower and upper quartile (includes interquartile range), lower and upper fence, outliers (marked with •) as well as the average value (marked with $\times$), for each aspect as recorded in our questionnaire.
  }
  \label{fig:userstudy:questions}
\end{figure*}
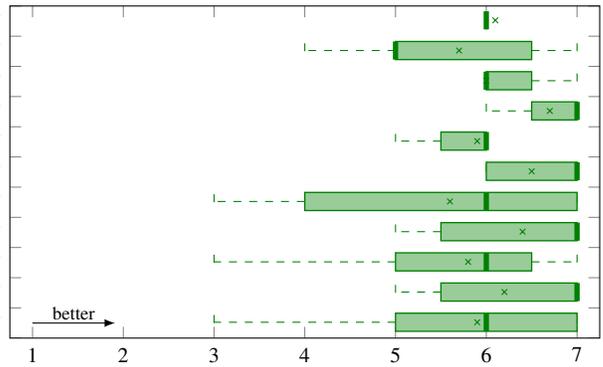

\subsection{User Study}

In the user study experiment, participants were asked to perform several
remote manipulation tasks. The avatar robot was stationed in a kitchen, out of sight
and hearing range of the operator (other end of the building).
The participants did not see the objects they were supposed to manipulate before
the task, although most of them were familiar with the kitchen.

Due to a motor failure of the right SVH hand, the robot was equipped with two
SIH hands for this test---slightly lowering the manipulation capabilities, as
the SVH hand is better suited for small-scale, dexterous manipulation.

See \cref{fig:userstudy:tasks} for an overview of the tasks to be performed.
For the first task, the avatar should navigate to three milk cartons.
The operators were told that one was full, one half full, and one empty.
They should use any means available to find out which was which.

The next three tasks were to put the full carton on a shelf, store the half-full
carton in the fridge (opening\footnote{During four runs, we experienced
hardware failures in the right hand and had to open the fridge with human assistance.}
and closing it), and throw the empty one into a waste basket.
These tasks could be performed in any order.
The fifth and final task was to move the waste basket from the kitchen counter to its proper place, requiring bimanual manipulation.

The ongoing COVID-19 pandemic limited us to our immediate colleagues
for the user study, severely constraining the scope of the study. Out of ten total participants, two were trained operators with deep knowledge of the system,
five more were members of the robotics group but were unfamiliar with the system,
and the final three were other staff members with no relation to robotics.
All untrained operators were given a few minutes to familiarize themselves with the
system, with the avatar being given a few unrelated example objects to manipulate.

\begin{table}
 \centering
 \caption{User study: Completion times and success rates}\label{tab:userstudy:times}\vspace{-.5em}
 \begin{tabular}{lrrr}
  \toprule
  Operator group & \multicolumn{2}{c}{Completion time [min:sec]} & Correct \\
  \cmidrule (lr){2-3}
                 & \hspace{2em} mean & stddev & \\
  \midrule
  Untrained (8) & 8:05 & 1:15 & 75\% \\
  Trained   (2) & 2:51 & 1:01 & 100\% \\
  \bottomrule
 \end{tabular}
\end{table}

\Cref{tab:userstudy:times} shows immediately quantifiable results. As a first
observation, it is apparent that the trained operators could perform the task three
to four times quicker, which is expected. Still, even untrained operators
could perform the task in very reasonable time, given the task complexity.
To put these times into perspective, our untrained operators solved the entire
mission with five tasks in 8 minutes, while the average task completion time of the top five teams during the DRC was around 5 minutes~\citep{schwarz2017nimbro}, excluding any locomotion---and with fully trained operators.
We feel that these results show that our operator interfaces are intuitive
and that the system can be immediately applied to challenging tasks by novices.
We did not notice any difference in performance between the robotics group members
and non-roboticists.

While all participants managed to solve the manipulation tasks, two operators
did not sort the milk cartons correctly, confusing the half-full and full cartons.
Nearly all operators relied on force feedback to judge the carton weight, with some
operators using visual feedback to confirm the empty carton, which responds differently
to contacts with the robot end-effector.

After performing the tasks, participants were asked to answer a questionnaire. We
again took inspiration from the ANA Avatar XPRIZE challenge rules, which specify
a series of questions for the human operator.
\Cref{fig:userstudy:questions} shows an aggregated view of the responses. We can
immediately see that the answers were mostly positive. Although there
was no baseline system available, we can gain insight by comparing the question responses to each other.

For example, nearly all operators felt completely present in the remote environment (Q4).
Furthermore, the operators felt safe while controlling the system (Q6), and said it
felt intuitive to operate the arms (Q8). On the downside, control of the fingers
was less intuitive, probably due to the inexact mapping between human and robot fingers
and the missing finger force feedback (as stated above, the more sensitive SVH hand was not available for the study).
Furthermore, the operators reported that they felt less sure about the safety on the
avatar side (Q7). When questioned, the participants indicated that this had mostly to
do with situational awareness during locomotion. While it is easy to see the space in
front of the robot, it is harder to see to the sides and impossible to see the space
behind the robot. We aim to improve this by adding separate rear cameras for locomotion
in the future.

\subsection{Integrated Mission}

\begin{figure}
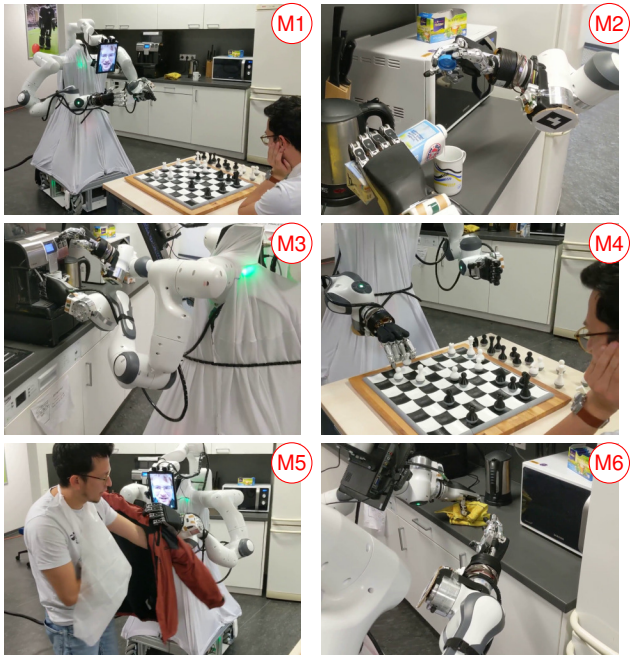

 \centering
 \tikzset{i/.style={inner sep=0cm,anchor=north east}, l/.style={inner sep=1pt,anchor=north east,draw=red,circle,text=red,font=\sffamily\footnotesize,fill=white,shift={(2pt,-2pt)}}}

 \tikz{\node[i]{\includegraphics[height=2.8cm,clip,trim=400 0 0 0]{images/mission/intro.png}}; \node[l]{M1}}
 \tikz{\node[i]{\includegraphics[height=2.8cm,clip,trim=400 0 0 0]{images/mission/milk.png}}; \node[l]{M2}}
 \\[0.3em]
 \tikz{\node[i]{\includegraphics[height=2.8cm,clip,trim=400 0 0 0]{images/mission/coffee.png}}; \node[l]{M3}}
 \tikz{\node[i]{\includegraphics[height=2.8cm,clip,trim=400 0 0 0]{images/mission/chess.png}}; \node[l]{M4}}
 \\[0.3em]
 \tikz{\node[i]{\includegraphics[height=2.8cm,clip,trim=200 0 200 0]{images/mission/jacket.png}}; \node[l]{M5}}
 \tikz{\node[i]{\includegraphics[height=2.8cm,clip,trim=0 0 400 0]{images/mission/clean.png}}; \node[l]{M6}}
 
 \vspace{-0.2cm}

 \caption{Integrated mission. The operator used the avatar to:
  1) Meet the recipient.
  2) Get some milk from the fridge.
  3) Make a coffee.
  4) Play some chess with the recipient.
  5) Help the recipient to put on his jacket.
  6) Clean up the kitchen.
 }
 \label{fig:mission:tasks}
\end{figure}

In the integrated mission, a trained operator performed a sequence of six tasks
involving locomotion, precise manipulation, and communication and interaction
with a recipient. The tasks are inspired by the ANA Avatar XPRIZE challenge rules
and designed to evaluate different system components (see \cref{fig:mission:tasks}).
In the first task, the operator meets the recipient (who suffers from an arm injury)
and offers help, specifically to make a coffee for the recipient. This demonstrates situational awareness (Where is the recipient?)
and good verbal and non-verbal communication.
The dexterous and precise manipulation capabilities are tested in the second and third tasks:
The avatar gets milk from the fridge, opens the carton, pours milk in the cup and
makes a coffee using the coffee machine. Especially opening and closing the fridge
requires a good haptic feedback for the operator.
The fourth and fifth tasks evaluate the human-robot interaction. First, the avatar and recipient
play a few moves of chess. This mainly tests vision capabilities (recognizing the
different chess pieces by shape) and precise manipulation capabilities (handling the
rather small chess pieces).
Afterwards, the operator helps the recipient
to put on his jacket, which involves close contact between the human and avatar. Good haptic feedback
and low impedance are key to make this experience non-threatening and comfortable
for the recipient.
Lastly, in the final task the avatar cleans up the kitchen which involves putting back
the milk into the fridge and wiping the kitchen counter. This task was designed to evaluate
the locomotion capabilities since a lot of different locations need to be accessed.
A video of this integrated mission is available\footnote{A video demonstrating the integrated mission can be found here:\\\scriptsize\url{http://ais.uni-bonn.de/videos/IROS_2021_NimbRo_Avatar}}.

\begin{table}
 \centering
 \caption{Integrated mission: Task completion times}\label{tab:integrated_mission:times}
 \small\vspace{-.5em}
 \setlength{\tabcolsep}{4pt}
 \begin{tabular}{l@{\hspace{12pt}}rrrrrrr}
  \toprule
  Trial & \multicolumn{7}{c}{Task completion time [s]}\\
  \cmidrule (l{0pt}r) {2-8}
                 & Intro & Milk & Coffee & Chess & Jacket & Clean & Total\\
  \midrule
  1 & 27 & 156 & 80  & 105 & 49 & 85  & 502 \\
  2 & 44 & 38  & -   & -   & -  & -   & - \\
  3 & 26 & 106 & 88  & 102 & 58 & 106 & 486 \\
  4 & 45 & 140 & 133 & 178 & 69 & 129 & 694 \\
  \bottomrule
 \end{tabular}
\end{table}

Although the operator was familiar with the system, most tasks were not trained beforehand.
We did test if the robot was able to open a milk carton before the trial.
\Cref{tab:integrated_mission:times} shows the timing results for the four attempted trials.
Considering that some tasks cannot be compared by time (i.e. conversations were not predefined and thus vary in length),
the results show quite similar task durations for all trials. Except for task two in Trial~2 (the avatar knocked over the milk inside the fridge), all tasks were executed successfully. The operator was able to correct smaller mistakes
such as dropping the lid of the milk carton onto the counter, or losing a chess piece by an imperfectly executed grasp, without
any external help.
Overall, this integrated mission experiment shows not only the raw capabilities of the robotic platform,
but also the reliability and intuitive control of the avatar which allow adaptation to unforeseen circumstances and tasks.

\subsection{Lessons Learned}

During both the user study and the integrated mission, we identified several
strong and weak points of our system, which may not be directly reflected in the
quantitative results already presented.
First of all, many user study participants expressed surprise at being back in their
actual environment, which they had forgotten about while controlling the Avatar.
This indicates that our system has a high level of immersion.

Second, some participants were significantly smaller than we had planned for,
resulting in difficulty reaching the 3D Rudder and, more significantly, resulting
in robot wrist poses closer to the body and thus more potential self-collisions.
These persons complained of jerky behavior as the system displayed forces in response
to position limit violations.
In a future iteration, we will address this by reducing the size of the avatar wrists
and by adapting the initial head pose to the operator size.

Overall, the system proved to operate reliably during all tests. We noticed that
the avatar arms stopped due to exceeded torque limits some times, but the automatic
recovery allowed safe continuation of operation. We will improve the control loop
further to avoid triggering the stop in any case.

\section{Conclusion}

We developed a system capable of executing complex dexterous and interaction tasks remotely.
It is especially suited for typical everyday indoor environments and interactions with both
remote made-for-human objects and environments as well as remote persons.
The system has proven itself in a small user study with untrained operators,
as well as in a longer and more complex integrated mission. The mission demonstrates
a chain of both highly difficult and useful tasks as well as
natural human-human interaction through the avatar.
We also identified some points of possible improvement, which we will address in future work.

\printbibliography

\end{document}